\documentclass[runningheads]{llncs}
\usepackage{graphicx}
\usepackage{amsmath, amssymb}
\usepackage{amssymb, amsmath}
\usepackage{amsmath}
\usepackage{type1cm}
\usepackage{enumerate}
\usepackage{algorithmic}
\usepackage{algorithm}

\begin{document}
\title{Robust Meta-Reinforcement Learning with Curriculum-Based Task Sampling}
\author{Morio Matsumoto \and
Hiroya Matsuba \and
Toshihiro Kujirai}
\authorrunning{M. Matsumoto et al.}
\institute{Research and Development Group, Hitachi, Ltd., Tokyo, Japan}
\maketitle              
\begin{abstract}
Meta-reinforcement learning (meta-RL) acquires meta-policies 
that show good performance for tasks in a wide task distribution.  
However, conventional meta-RL, 
which learns meta-policies by randomly sampling tasks, 
has been reported to show meta-overfitting for certain tasks, 
especially for easy tasks where an agent can easily get high scores.
To reduce effects of the meta-overfitting, 
we considered meta-RL with curriculum-based task sampling. 
Our method is Robust Meta Reinforcement Learning with Guided Task Sampling (RMRL-GTS), 
which is an effective method that restricts task sampling 
based on scores and epochs. 
We show that in order to achieve robust meta-RL, 
it is necessary not only to intensively sample tasks with poor scores, 
but also to restrict and expand the task regions of the tasks to be sampled.

\end{abstract}

%%%%%%%%%%%%%%%%%%%%%%%%%%%%%%%%%%%%%%%%%%%%%%%%%%%%%%%%%%%%%%%%
%  Outline
%%%%%%%%%%%%%%%%%%%%%%%%%%%%%%%%%%%%%%%%%%%%%%%%%%%%%%%%%%%%%%%%%

\section{Introduction}
Reinforcement learning (RL) has been expected to be applied to real-world tasks in various industries.
An RL agent  learns policies for a given task through many trials. 
After training, RL agent showed scores that exceeded human average scores \cite{Badia} in Atari and 
RL agents have acquired good performance in the field of 
robotics \cite{Kober}. 
However, agents may not learn good policies for real-world's tasks due to model errors 
between the simulation and the real-world. 
Model errors are difficult to eliminate completely, 
even if the simulation is created accurately, due to environmental changes.

It is necessary to learn robust policies in order to apply RL to real-world tasks 
by using simulation with model errors. 
RL with Domain Randomization enables an RL agent to 
adapt to various tasks with learned policies in various simulation situations. 
This method can be used to get a useful policy  
even in the real world \cite{James}. 
This indicates that an RL agent has to learn a robust policy to adapt to a task region that corresponds to various simulation situations. 
However, the results with Domain Randomization do not show a $100\%$ success rate. 
This implies 
that the real-world tasks might be more difficult than train tasks that an RL agent learns to acquire a robust policy. 
In addition, Domain Randomization is difficult to adapt to failed tasks in the real world with a few trials of learning, 
so that it takes a lot of time to relearn the policy.  
Thus, robust methods are required to adapt to real-world tasks which are more difficult tasks than 
train tasks with a few trials of learning. 

Meta-RL has been expected to be one of robust RL methods. 
Model-Agnostic-Meta-Leaning (MAML) \cite{Finn} is well known as a popular meta-RL method. 
MAML makes an RL agent acquire a meta policy  by sampling tasks from task distribution, 
which corresponds to various simulation situations.  
\cite{Finn} reported that a meta policy acquired with MAML could adapt to various tasks 
sampled from a wide task distribution with a few trials of learning. 
However, \cite{Mehta} reported that MAML shows 
a low score for some tasks when the task distribution is wider than the distribution considered in \cite{Finn}. 
In addition, they also reported that MAML overfits specific tasks, especially easy tasks, in which 
a score after adaptation is high \cite{Mehta}. 
These reports indicate that a meta-policy  
cannot adapt 
to tasks sampled from a wide task distribution. 
As discussed in \cite{Hospe}, 
meta-learning results are not a good performance for simulation with a wide task distribution 
and for test tasks that are not learned during meta-training. 
The problem of meta-overfitting is the same as this open problem. 
Although improved approaches such as probabilistic embeddings for actor-critic RL 
(PEARL) \cite{Rakelly} improve scores, there is still some bias 
in task scores for the task distribution,
 and these approaches show low scores for difficult tasks sampled from wide 
task distributions \cite{Lin}. 
This indicates a new method is required to make meta-policies more robust 
and to be adapted to tasks in a wide task distribution.

In this paper, 
we propose Robust Meta-Reinforcement Learning 
with Guided Task Sampling (RMRL-GTS) to decrease the effects of meta-overfitting 
with curriculum learning. 
Since meta-learning which randomly samples tasks such as MAML 
does not consider the difficulty of the task, 
the opportunity to learn difficult tasks is insufficient, and the meta-policy cannot adapt to the difficult tasks. 
Thus, our method changes the sampling rates of tasks by using curriculum 
that restricts task sampling based on scores and the sampling region. 
We show that both sampling of tasks with scores and restricting the task region to be sampled is necessary to achieve robust meta-RL.

\section{Related Works}
To solve the problem of meta-overfitting as seen in methods such as 
MAML \cite{Finn} and PEARL \cite{Rakelly}, 
\cite{Mehta} considered meta-ADR, which learns tasks  
using MAML and curricula: Active Domain Randomization (ADR) \cite{Mehta2}. 
A feature of meta-ADR is that it uses sampling by curriculum learning using ADR 
instead of random sampling used in MAML. 
This feature enables 
an RL agent to concentrate on learning tasks where the agent's trajectory before and after adaptation are 
different. 
\cite{Mehta} reported that MAML shows  meta-overfitting in various meta RL benchmarks. 
On the other hand, meta-ADR shows less score bias in some meta RL benchmarks, 
which indicates it is a robust method in meta-learning. 
However, 
meta-ADR scores were worse than MAML on the Ant-Velocity task 
in MuJoCo \cite{Todorov}, especially on difficult tasks 
(large target velocity). 
This task is more difficult than other meta-benchmarks where meta-ADR 
scored better than MAML since the agent must acquire the skill to run 
at a faster velocity as the velocity sampled from the task distribution increases.

Another approach, Model-based Adversarial Meta-Reinforcement Learning (AdMRL) \cite{Lin}, 
shows high scores in a wide range of task distributions 
in various simulations including Ant-Velocity. 
AdMRL uses the information on the reward function gradient with respect to 
the task parameters, such as target velocity. 
This indicates that AdMRL cannot be applied to a simulation where the task 
information on the reward function is unclear. 
In real-world tasks, the relation between 
the reward functions and task distributions is often unknown.

Our basic idea is similar to \cite{Mehta}, which changes the sampling method 
by curriculum learning. Our proposed method, RMRL-GTS, 
changes sampling rates of tasks in a task distribution during meta-learning 
on the basis of the score and epoch information obtained at each epoch 
during meta-training. 
Although our proposed curriculum learning is quite simple, 
RMRL-GTS shows better scores than MAML for difficult tasks.

\section{Preliminaries}

We explain to reinforcement learning and 
gradient-based meta-RL, 
MAML. 

\subsection{Reinforcement Learning}
We consider 
Markov decision process (MDP) 
$M(\pi) := \{ \mathcal{S}, \mathcal{A}, p_0, p_T, \mathcal{R}, \pi \} $. 
$\mathcal{S}$ is a state space, $\mathcal{A}$ is an action space. 
$p_0: \mathcal{S} \rightarrow [0, 1]$ is an initial state distribution 
$p_{0}(s_{t=0})$, and 
$p_T: \mathcal{S}\times \mathcal{S}\times \mathcal{A} \rightarrow [0, 1] $ 
is a transition probability $p_T(s_{t+1}| s_t, a_t)$, 
where $t$ is a time step. $\mathcal{R} := \{r\in \mathbb{R}: r = g(s_t, a_t), \exists (s_t, a_t) \in \mathcal{S} \times \mathcal{A} \}$ 
is a reward function at each time step $t$. 
$\pi: \mathcal{A}\times \mathcal{S}\rightarrow [0, 1]: \pi(a_t|s_t)$ 
is a policy, which determines the action with a state. 
The goal of RL is to learn $\pi$, which maximizes the total expected 
discounted reward with a discount rate $\gamma$. 
Policy gradient methods with deep neural networks such as Trust Region Policy Optimization (TRPO) \cite{Schulman} 
is a type of RL method, 
and $\pi$ is parameterized as $\pi_\theta$ with $\theta$. 
TRPO takes the gradient of policy $\pi$ with respect to the neural network parameter $\theta$.

\subsection{Gradient-based Meta-Reinforcement Learning}
In this paper, we consider a popular meta-RL, 
gradient-based meta-learning algorithm, MAML \cite{Finn} with TRPO. 
The goal of MAML is to a learn meta-policy $\pi_\theta$ with the initial neural network parameter $\theta$ (meta-parameter), 
which can quickly adapt to new task $\tau$ sampled from task distribution $p(\tau)$. 
Here, $\tau$ is a parameter that changes simulation situation in which an RL agent learns. 
For example, in Ant-Velocity, $\tau$ corresponds to target velocity (see Sec. \ref{sec:experiment_settings}). 
To acquire a meta-policy $\pi_\theta$, MAML takes a few gradient steps. 
First, MAML calculates parameters $\theta_i$, 
which is optimized for task $\tau_i$ written as

\begin{equation}
\theta'_i = \theta - \alpha \nabla_\theta \mathcal{L}_{\tau_i}(\pi_{\theta}). \label{eq:pre-meta-1}
\end{equation}
Here, $\mathcal{L}_{\tau_i}$ is a loss function of a task $\tau_i$, 
$\alpha$ is the step size. 
$N_{\rm meta}$ tasks are randomly sampled from $p(\tau)$ and 
each $\theta'_i$ is calculated with Eq. (\ref{eq:pre-meta-1}). 
Meta-parameter $\theta$ is then updated as 
\begin{equation}
\theta = \theta - \beta \nabla_\theta \sum_{\tau_i \sim p(\tau)}\mathcal{L}_{\tau_i}(\pi_{\theta'_i}), \label{eq:pre-meta-2}
\end{equation}
where $\beta$ is the step size. In this paper, the process of Eq. (\ref{eq:pre-meta-1}) is defined as adaptation. 
The process of Eq. (\ref{eq:pre-meta-2}) including Eq. (\ref{eq:pre-meta-1}) is repeated $N_{\rm epoch}$ times.

%%%%%%%%%%%%%%%%%%%%%%%%%%%%%%%%%%%%%%%%%%%%%%%%%%%%%%%%%%%%%%%%%%%%

\section{Proposed Method}

\begin{figure}[!t]
  \begin{algorithm}[H]
      \caption{RMRL-GTS}
      \label{alg1}
      \begin{algorithmic}[1]
      \REQUIRE $p(\tau)$: task distribution
      \REQUIRE $N_{\rm epoch}, N_{\rm sample}, n_{\rm interval}, \tau_{\rm min}, \tau_{\rm max}$
      \STATE score: $R_{\rm mean}(\tau)$ for all tasks $\tau$ in $p(\tau)$ are initialized to $0$
      \STATE $n_{\rm ce} = 0$, task regions: $T_{\rm easy}, T_{\rm middle}, T_{\rm difficult}$ are None
      \WHILE{$n_{\rm ce} \neq N_{\rm epoch}$}
      \IF{$n_{\rm ce}>0$}
      \STATE  $T_{\rm easy}, T_{\rm middle}, T_{\rm difficult}\leftarrow$ApproachI($R_{\rm mean}(\tau)$, $\tau_{\rm min}, \tau_{\rm max}$, $N_{\rm epoch}, n_{\rm ce}, n_{\rm interval}$)
      \ENDIF
      \STATE $\tau$$\leftarrow$ApproachII$(n_{\rm ce}, N_{\rm sample}, \tau_{\rm min}, \tau_{\rm max}, d\tau_{\rm bin}, R_{\rm mean}(\tau), T_{\rm easy}, T_{\rm middle},T_{\rm difficult})$
      \STATE $R_{\rm mean}(\tau)$  $\leftarrow$ MAML($\tau$)
      \IF{$n_{\rm ce}=0$}
      \STATE $T_{\rm easy}, T_{\rm middle}, T_{\rm difficult}\leftarrow$ApproachI($R_{\rm mean}(\tau)$, $\tau_{\rm min}, \tau_{\rm max}$, $N_{\rm epoch}, n_{\rm ce}, n_{\rm interval}$)
      \ENDIF
      \STATE $n_{\rm ce}\leftarrow n_{\rm ce} +1$
      \ENDWHILE
      \end{algorithmic}
  \end{algorithm}
  \end{figure}

In this section, we introduce our proposed method, RMRL-GTS. 
The overall algorithm of the RMRL-GTS is described in Algorithm \ref{alg1}.
As shown in \cite{Mehta}, MAML and their approach meta-ADR 
tend to cause meta-overfitting 
to easier tasks in task distribution $p(\tau)$ for meta-RL benchmarks 
such as Ant-Velocity. 
Thus, our method, RMRL-GTS, aims to suppress meta-overfitting 
by improving approaches of sampling tasks. 
Our method consists of two simple approaches:

\begin{enumerate}[I]
\item :  Restriction of the task sampling region 
\item :  Prioritized sampling with a score
\end{enumerate}

\subsection{Approach I: Restriction of task sampling region}\label{sec:approach1}
\begin{figure}[!t]
  \begin{algorithm}[H]
      \caption{Approach I}
      \label{alg2}
      \begin{algorithmic}[1]
      \REQUIRE $N_{\rm epoch}, n_{\rm ce},  n_{\rm interval}, \tau_{\rm min}, \tau_{\rm max}$, and a list of scores
      \IF{current epoch number $n_{\rm ce}=0$}
      \STATE $\tau_{\rm mean}$ is determined with a list of scores
      \STATE $n_{\rm batch}=0.5N_{\rm epoch}/n_{\rm interval}$
      \STATE $d\tau = (\tau_{\rm max} - \tau_{\rm mean})/n_{\rm interval}$
      \STATE $\tau_{\rm middle1}\leftarrow \tau_{\rm mean}$
      \STATE $\tau_{\rm middle2}\leftarrow \tau_{\rm mean} + 0.5 * (0.5N_{\rm epoch}/n_{\rm batch})*(0.5d\tau)$
      \ELSIF{$n_{\rm ce} > 0.5 N_{\rm epoch}$}
      \IF{mod($n_{\rm ce}-0.5N_{\rm epoch},n_{\rm batch}$)}
      \STATE $\tau_{\rm middle1}\leftarrow \tau_{\rm middle2}$
      \STATE $\tau_{\rm middle2}\leftarrow \tau_{\rm middle2} + d\tau$
      \ENDIF 
      \ENDIF
      \STATE $T_{\rm easy}, T_{\rm middle}, T_{\rm difficult}$ are determined with $\tau_{\rm middle1}, \tau_{\rm middle2}, \tau_{\rm min}, \tau_{\rm max}$
      \end{algorithmic}
  \end{algorithm}
  \end{figure}
\begin{figure}[t]
  \includegraphics[width=\textwidth]{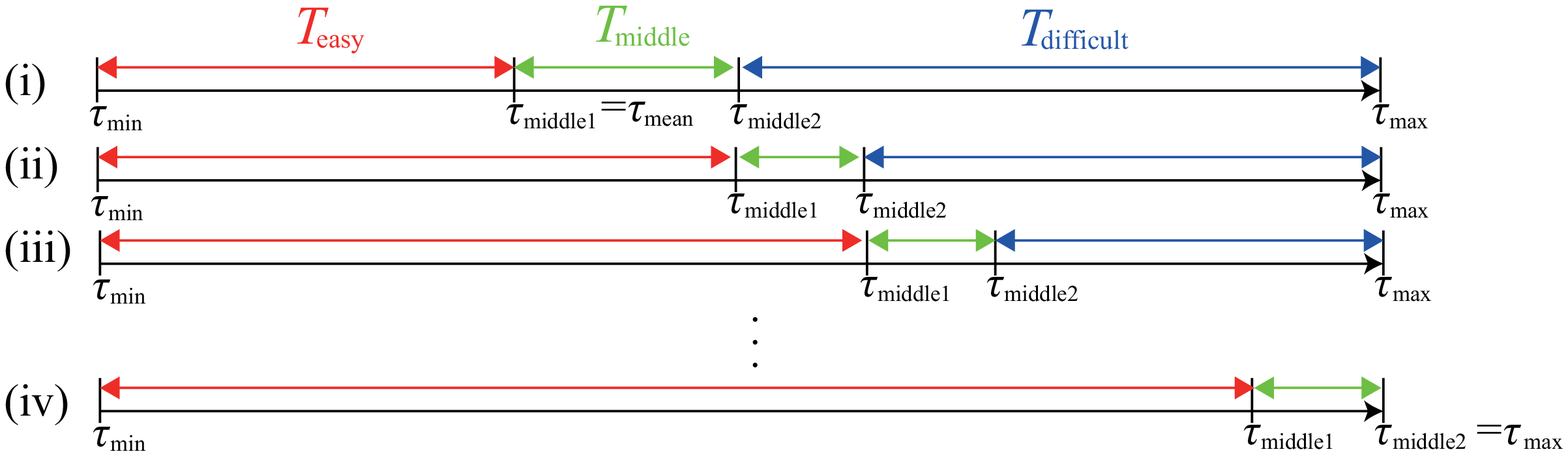}
  \caption{Overview of approach I for 1D task distribution and $\tau_{\rm min}\geq 0$. 
            $T_{\rm easy}$, $T_{\rm middle}$, $T_{\rm difficult}$ are changed as epoch increases. 
            (i) $0\leq$ epoch$\leq 0.5N_{\rm epoch}$, (ii) $0.5N_{\rm epoch}\leq$epoch$\leq 0.5N_{\rm epoch}+n_{\rm batch}$, 
            (iii) $0.5N_{\rm epoch}+n_{\rm batch}\leq$epoch$\leq 0.5N_{\rm epoch}+2n_{\rm batch}$, 
            and (iv) $0.5N_{\rm epoch}+(n_{\rm interval}-1)n_{\rm batch}\leq$epoch$\leq N_{\rm epoch}$.} \label{fig:approach1}
\end{figure}

Approach I restricts the task sampling region with epochs 
and task difficulty as seen in Algorithm \ref{alg2} and 
Fig.\ref{fig:approach1}. 
For the sake of simplicity, at first, 
we consider a simple 1D task distribution 
$\tau \sim p(\tau), 0\leq \tau_{\rm min}\leq \tau \leq \tau_{\rm max}$. 
Note that Approach I can be applied to the case with a task distribution 
including negative $\tau$ and a 2D task distribution 
(See Eq. (\ref{eq:aprroach1-negative}) and Sec. \ref{sec:experiment_settings}). 
In addition, 
we assume that the value of $\tau$ for a task is bigger, the more difficult the task is. 
We do not need the previous assumption when we transform 
the coordinates of task parameter $\tau$ to new coordinates of task parameter $\tau_{\rm new}$ at epoch $=0$ 
to satisfy that the bigger $\tau_{\rm new}$ is, the more difficult task is. 

Our method determines the initial difficulty of tasks 
at the first epoch. 
At epoch $=0$, tasks are sampled 
at constant intervals between $\tau_{\rm min}$ and $\tau_{\rm max}$.  
The number of the intervals corresponds to the number of sampling tasks $N_{\rm batch}$ in an epoch. 
The total mean reward is then determined by summing total rewards of each task, 
and the task with the total reward closest to the total mean reward is defined as $\tau_{\rm mean}$. 
We define three tasks regions, 
$T_{\rm easy}$, $T_{\rm middle}$, and $T_{\rm difficult}$ as follows. 

\begin{align}
  &T_{\rm easy}: \tau_{\rm min} \leq \tau < \tau_{\rm middle1} \notag \\
  &T_{\rm middle}: \tau_{\rm middle1} \leq \tau < \tau_{\rm middle2} \label{eq:aprroach1-positive} \\
  &T_{\rm difficult}: \tau_{\rm middle2} \leq \tau \leq \tau_{\rm max} \notag 
\end{align}
At epoch $=0$, $\tau_{\rm middle1}=\tau_{\rm mean}$, and 
$\tau_{\rm middle2} = \tau_{\rm mean} + 0.5 * (0.5N_{\rm epoch}/n_{\rm batch})$ $* (0.5  d\tau)$. 
We define $n_{\rm batch} = 0.5  N_{\rm epoch}/n_{\rm interval}$. 
Here, $N_{\rm epoch}$ is total epoch size, 
the task region changes every $n_{\rm batch}$ after the latter epoch 
($0.5  N_{\rm epoch}$), 
and $n_{\rm interval}$ 
is the number of times the task region is changed. 
$d\tau$ is written as $d\tau= (\tau_{\rm max} - \tau_{\rm mean})/n_{\rm interval}$. 

Tasks are sampled from $T_{\rm easy}$ and $T_{\rm middle}$, 
which are fixed until 
epoch $=0.5N_{\rm epoch}$. 
$0.5N_{\rm batch}$ tasks are chosen from 
$T_{\rm easy}$ and other tasks are chosen from $T_{\rm middle}$. 
$T_{\rm easy}, T_{\rm middle}$ and $T_{\rm difficult}$ are updated after epoch 
$\geq 0.5 N_{\rm epoch}$. 
At each $n_{\rm batch}$ after epoch $\geq 0.5 N_{\rm epoch}$, 
$\tau_{\rm middle1} \leftarrow \tau_{\rm middle2}$ and 
$\tau_{\rm middle2} \leftarrow \tau_{\rm middle2} + d\tau$. 
When ${\rm{mod}}({\rm epoch}-0.5N_{\rm epoch}, n_{\rm batch})=0$, 
$T_{\rm easy}$ expands, 
$T_{\rm middle}$:$d\tau$, and $T_{\rm difficult}$ shrinks.  
Finally, $T_{\rm easy}+T_{\rm middle}$: $\tau_{\rm min} \leq \tau \leq \tau_{\rm max}$ as seen in Fig. \ref{fig:approach1}.

At first, the aforementioned task region is $0\leq \tau_{\rm min}\leq \tau \leq \tau_{\rm max}$. 
When $\tau_{\rm min} < 0$ and $\tau_{\rm min} = - \tau_{\rm max}$, 
the task region is defined as follows.
\begin{align}
  &T_{\rm easy}: -\tau_{\rm middle1} \leq \tau < \tau_{\rm middle1} \notag \\
  &T_{\rm middle}: -\tau_{\rm middle2} < \tau \leq -\tau_{\rm middle1}, \tau_{\rm middle1} \leq \tau < \tau_{\rm middle2}  \label{eq:aprroach1-negative} \\
  &T_{\rm difficult}: \tau_{\rm min} \leq \tau \leq -\tau_{\rm middle2}, \tau_{\rm middle2} \leq \tau \leq \tau_{\rm max} \notag
\end{align}
$0.5*N_{\rm batch}$ tasks are chosen from $T_{\rm easy}$ and other tasks are 
chosen equally from the positive and negative task regions in $T_{\rm middle}$.

\subsection{Approach II: Prioritized sampling with score}\label{sec:approach2} 
\begin{figure}[!t]
  \begin{algorithm}[H]
      \caption{ Approach II}
      \label{alg3}
      \begin{algorithmic}[1]
      \REQUIRE $n_{\rm ce}, N_{\rm sample}, \tau_{\rm min}, \tau_{\rm max}, d\tau_{\rm bin}$, a list of scores, $T_{\rm easy}, T_{\rm middle}, T_{\rm difficult}$
      \IF{current epoch number $n_{\rm ce}=0$}
      \STATE $N_{\rm sample}$ tasks are sampled at constant intervals between $\tau_{\rm min}$ and $\tau_{\rm max}$
      \ELSE
      \STATE Tasks and scores are grouped at intervals of $d\tau_{\rm bin}$ (binning)
      \STATE Scores are calculated with Eq.(\ref{eq:method2-1})
      \STATE Interpolated function $\bar{f}$ is created with scores
      \STATE Interpolated scores $\leftarrow \bar{f}($scores$, d\tau_{\rm bin})$
      \STATE Task distribution $p(\tau)$ is calculated with Eq. (\ref{eq:method2-3})
      \STATE $N_{\rm sample}$ tasks are sampled with $p(\tau)$
      \ENDIF
      \end{algorithmic}
  \end{algorithm}
\end{figure}
Approach II samples difficult tasks in 
preference to easy tasks as seen in Algorithm \ref{alg3}.
The mean total reward is calculated with a weight that considers 
the current epoch number, and we convert the reward to probability. 
We define the mean total reward (score) with a weight for the epoch number as
\begin{equation}
f(\tau) = \frac{1}{n_{\rm bin}}\sum_{\tau<\tau'<\tau+d\tau_{\rm bin}} \sum_{i=0}^{n_{\rm ce}} c_i(\tau') R_{\rm mean, i}(\tau'). \label{eq:method2-1}
\end{equation}
Here, $c_i$ is the weight with the epoch written as 
  \begin{equation}
  c_i(\tau) = 
  \begin{cases}
  i \quad(\tau {\rm \, is\, sampled\, at\, epoch\,} i)\\
  0 \quad({\rm others}). \label{eq:method2-2}
  \end{cases}
\end{equation}
$c_i(\tau)$ indicates that the mean total reward with a large number of 
epochs is more important in $f(\tau)$. 
$n_{\rm ce}$ is the current epoch number, 
and $R_{{\rm mean,} i}(\tau)$ is the mean total reward function $R_{\rm mean}(\tau)$ 
at epoch$=i$ with reward function $r_j(t; \tau)$ at each time step $t$ 
at task $\tau$ with sample $j$ written as 
\begin{equation}
  R_{\rm mean}(\tau) = \frac{1}{N_{\rm samples}} \sum_{j=0}^{N_{\rm samples}} \sum_{t=0}^{t=H}r_j(t; \tau), \label{eq:reward-total}
\end{equation}
where $H$ is the horizon of an episode.
In addition,  
$R_{\rm mean}(\tau)$ 
is also averaged with 
the number of trial $N_{\rm samples}$ for a task. 
Furthermore, when $R_{\rm mean}(\tau)$ is calculated, 
a certain task region is averaged 
and 
considered as one task (binning) 
to consider discrete probability distribution $p(\tau)$ 
to reduce the effects of score variability. 
We define the binning interval as $d\tau_{\rm bin}$ and $n_{\rm bin}$ is the number of data in the bin of each task.

To calculate the probability with $f(\tau)$, 
the mean total rewards for no sampled tasks need to be determined. 
Thus, our method interpolates $f(\tau)$ with observed scores. 
In this paper, we use spline interpolation as one of the interpolation methods. 
In addition, our method normalizes $f(\tau)$ with min-max normalization. 
We define the interpolated normalized $f(\tau)$ as $\bar{f}(\tau)$.

The probability is defined with the interpolated normalized $\bar{f}(\tau)$ as 
\begin{equation}
p(\tau) = \frac{1-\bar{f}(\tau)}{\sum_{\tau}1-\bar{f}(\tau)}. \label{eq:method2-3}
\end{equation}
Here, since the task interval sampled from $p(\tau)$ is constant $d\tau_{\rm bin}$, 
it gives a random noise to the value of each $\tau$. 
When $\bar{f}(\tau)$ for a task $\tau$ is small, in other words, a low mean total reward 
(a difficult task), 
Eq. (\ref{eq:method2-3}) shows the task 
$\tau$ can be sampled a lot. 
If a task $\tau$ is not sampled, the score at current epoch 
is not updated in $\bar{f}(\tau)$, 
so that it is possible that although $\bar{f}(\tau)$ is large when the epoch is small, 
$\bar{f}(\tau)$  including the score at the current epoch is small. 
$\bar{f}(\tau)$ at small $\tau$ in $T_{\rm easy}$ tends to be high when the epoch is small 
since the task is easy at epoch$=0$. 
Thus, Approach II randomly samples tasks at a constant rate 
$\delta$ in easy task region $T_{\rm easy}$ 
so that the scores for tasks in $T_{\rm easy}$ do not decrease. 
We note that Approach II samples tasks with Eq. (\ref{eq:method2-3}) at a rate $1-\delta$ in $T_{\rm easy}$. 
Approach II corresponds to completely random sampling in $T_{\rm easy}$ when $\delta=1$.

\section{Results}

We evaluated our method for Ant-Velocity-task in MuJoCo \cite{Todorov}  
and 2D Navigation \cite{Finn}. 
We described experimental settings in Sec. \ref{sec:experiment_settings}. 
In addition, we evaluated our method for HalfCheetah-Velocity-task in MuJoCo in Sec. \ref{sec:halfcheetah}. 
As in the case of the Ant-VelocityTask, the score is higher for tasks where the proposed method is difficult.

\subsection{Experimental settings}\label{sec:experiment_settings}
In MuJoCo, we evaluated the performance with Ant-Velocity and 
HalfCheetah-Velocity with an agent (Ant-V2 and HalfCheetah-V2). 
These are standard meta-RL benchmarks. 
Ant with four legs has 111 state dimensions and 8 action dimensions. 
HalfCheetah with two legs has 17 state dimensions and 6 action dimensions. 
These action spaces are continuous values. 
The goal of the velocity tasks is that the agent runs at a specific velocity 
chosen from task distribution $v_{\rm target}\sim p(\tau)$. 
The information on the target velocity is not given to the agent directly, 
however, the agent can know the target velocity from changes in the reward 
function. The reward function is described as 
\begin{equation}
 r(v; v_{\rm target}) := -|v - v_{\rm target}| + 1.0 + r_{\rm survive} + r_{\rm action}. \label{eq:reward-vel}
\end{equation}
Here, $r_{\rm survive} = 0.05$, which is given until the agent fails to run, 
and $r_{\rm action}$ is the cost for control whose detail is defined in 
\cite{Deleu}. These rewards' contributions are smaller than other terms 
($-|v - v_{\rm target}| + 1.0$) in Eq. (\ref{eq:reward-vel}). 
The agent acquires this reward function at each time step $t$.
 We evaluated whether our method is better than MAML or a previous method \cite{Mehta} 
 with the score being the mean total reward defined in Eq. (\ref{eq:reward-total}). 

2D Navigation is also a standard meta-RL benchmark. 
In this benchmark, a point agent must move from $(0,0)$ to goal positions chosen from 
task distribution $ (x_{\rm goal}, y_{\rm goal}) \sim p(\tau)$. 
The observations are the agent's current 2D position (x, y), the action 
is continuous clipped velocity commands are in the range 
$[-0, 1, 0.1]$. 
The reward function at time step $t$ is a negative value of the distance from the goal 
position to the agent's current position written as 
\begin{equation}
 r(x, y; x_{\rm goal}, y_{\rm goal}) := -\sqrt{(x-x_{\rm goal})^2 + (y-y_{\rm goal})^2}. \label{eq:reward-2d}
\end{equation}

All hyper-parameters for each benchmark were the same as those in \cite{Finn,Mehta} 
except the meta-training epoch $N_{\rm epoch}$. 
In addition, we used the final policy at the end of $N_{\rm epoch}$ instead of 
the best meta-policy over $N_{\rm epoch}$ as with the previous research \cite{Mehta}. 
Furthermore, we also used the source code \cite{Deleu} 
in our experiments, which was also the same as \cite{Mehta}. 
We averaged all test results with five seed results, 
and the results were the scores after adaptation to each task.

The hyper-parameters in our method were as follows. 
$N_{\rm epoch}$:$[200, 1000]$, ($N_{\rm epoch} = 200$ in 2D Navigation), 
$N_{\rm batch} = 40$ (MuJoCo), $N_{\rm batch} =20$ (2D Navigation), 
$N_{\rm samples} = 20$, $n_{\rm interval} = 10$, $d\tau_{\rm bin} = 0.1$, 
$\delta=0.1$, horizon $H = 100$, 
and the task interval for each test was $0.05$.

\subsection{MuJoCo: Ant-Velocity}

In the Ant-Velocity task, \cite{Mehta} shows that their approach is not 
superior to MAML, especially in the difficult task region: $v \gtrsim 2$. 
We show that the meta-policy acquired by RMRL-GTS can be adapted to 
more difficult tasks than MAML.

In our experiments, the region of the distribution is $ v: [0, 3]$, 
which is in common with that of MAML and \cite{Mehta}. 
Note that we only evaluated the score in the positive velocity region 
although \cite{Mehta} evaluated scores even in the negative velocity region. 
This is because it is appropriate to include the negative velocity 
in task distribution if we evaluate the score with negative target velocity. 
In the velocity task, 
the task region with positive velocity is often discussed, 
so we only evaluated the score with positive target velocity. 
In Sec. \ref{sec:2DNavi}, 
we evaluated the positive and negative regions 
of task distribution in 2D Navigation.

%%%%%%%%%%%%%%%%%%%%%%%%%%%%%%%%%%%%%%%%%%%%%%%%%%%
%%%% MAML VS Our Method at N_epoch = 200
%%%%%%%%%%%%%%%%%%%%%%%%%%%%%%%%%%%%%%%%%%%%%%%%%%%

\begin{figure}[!t]
  \includegraphics[width=\textwidth]{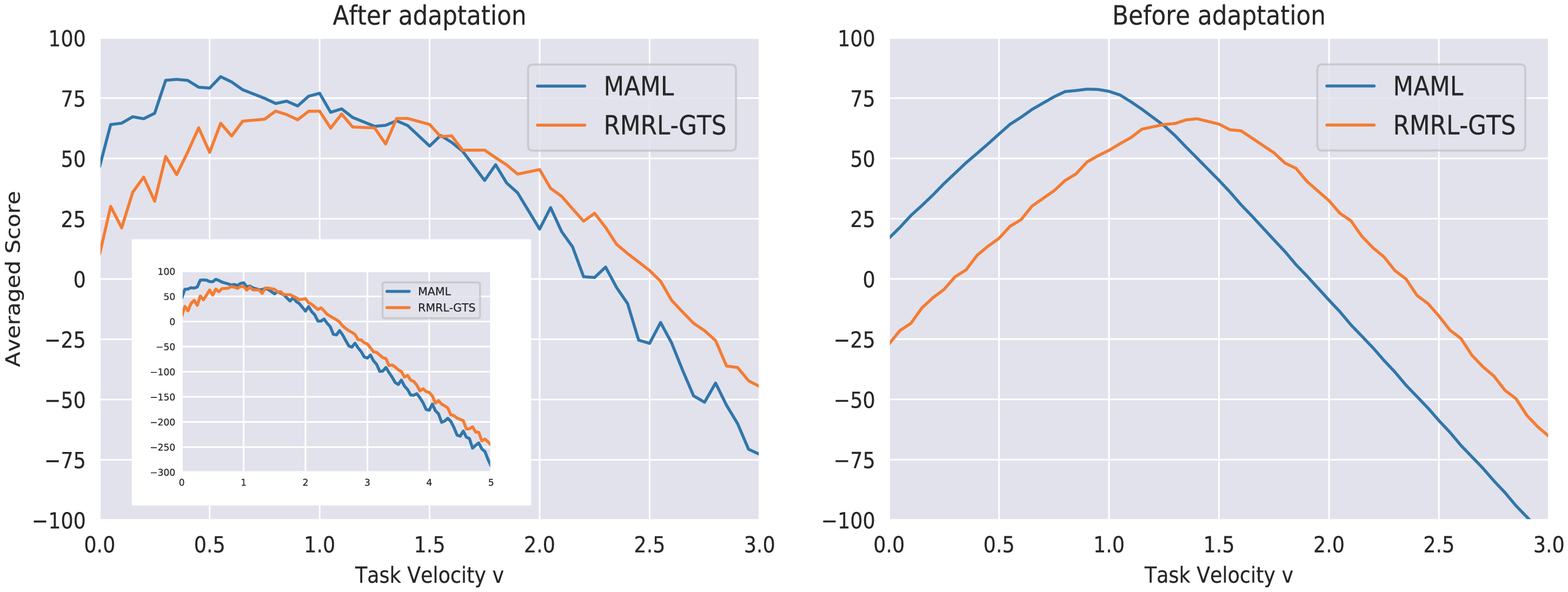}
  \caption{Averaged scores with five seeds for test tasks after and 
  before adaptation by MAML and RMRL-GTS for Ant-Vel Task 
  at $N_{\rm epoch}=200$. 
          } \label{fig:ant_epoch_200}
\end{figure}

Figure \ref{fig:ant_epoch_200} shows the scores after and before adaptation 
to each task. Here, adaptation means gradient step in MAML 
(Eq. (\ref{eq:pre-meta-1})). 
Note that the meta-policy was reset to pre-adaption policy each time it was applied to a task. 
MAML shows better-averaged scores with five seeds than RMRL-GTS 
for easier tasks ($v \lesssim 1.5$). However, for more difficult tasks 
($v \gtrsim 1.5$),  RMRL-GTS acquires higher scores than MAML. 
In addition, with our method, the score differences in the meta-training 
task region ($v: [0,3]$) are relatively small, in other words, 
the difference between the max and low scores is smaller than that of MAML. 
This result implies that MAML shows meta-overfitting to easy tasks 
($v: [0,1]$) where MAML's scores are higher than those acquired in $v\gtrsim 1$. 
Our method decreases the bias of scores in the task distribution and 
also shows a higher score than MAML even out of task the distribution ($v: [3,5]$). 
This result indicates that RMRL-GTS can adapt to more difficult tasks 
and is more robust than MAML. 
Furthermore, our method is more robust than that of \cite{Mehta}, 
in which it was shown to have lower scores than those of MAML in the 
Ant-Velocity task.

Our method is shown to be more robust than MAML even in the right figure 
of Fig. \ref{fig:ant_epoch_200}. The results show the scores with the 
meta-policy before adaptation to each task, i.e. the scores before the policy 
is not updated by Eq. (\ref{eq:pre-meta-1}). 
The highest score acquired by our method is for a task with a higher velocity 
than those with MAML, and especially, this velocity is near the middle of 
the task distribution. 
This result implies our method can be applied to more difficult tasks 
than MAML in the task distribution.

%%%%%%%%%%%%%%%%%%%%%%%%%%%%%%%%%%%%%%%%%%%%%%%%%%%%%%%%%%%%%%%%%%%%%
%%% Evaluation Sampling efficiency in Our Method at N_epoch = 200
%%%%%%%%%%%%%%%%%%%%%%%%%%%%%%%%%%%%%%%%%%%%%%%%%%%%%%%%%%%%%%%%%%%%%
\begin{figure}[!t]
  \begin{center}
 \includegraphics[width=0.8\textwidth]{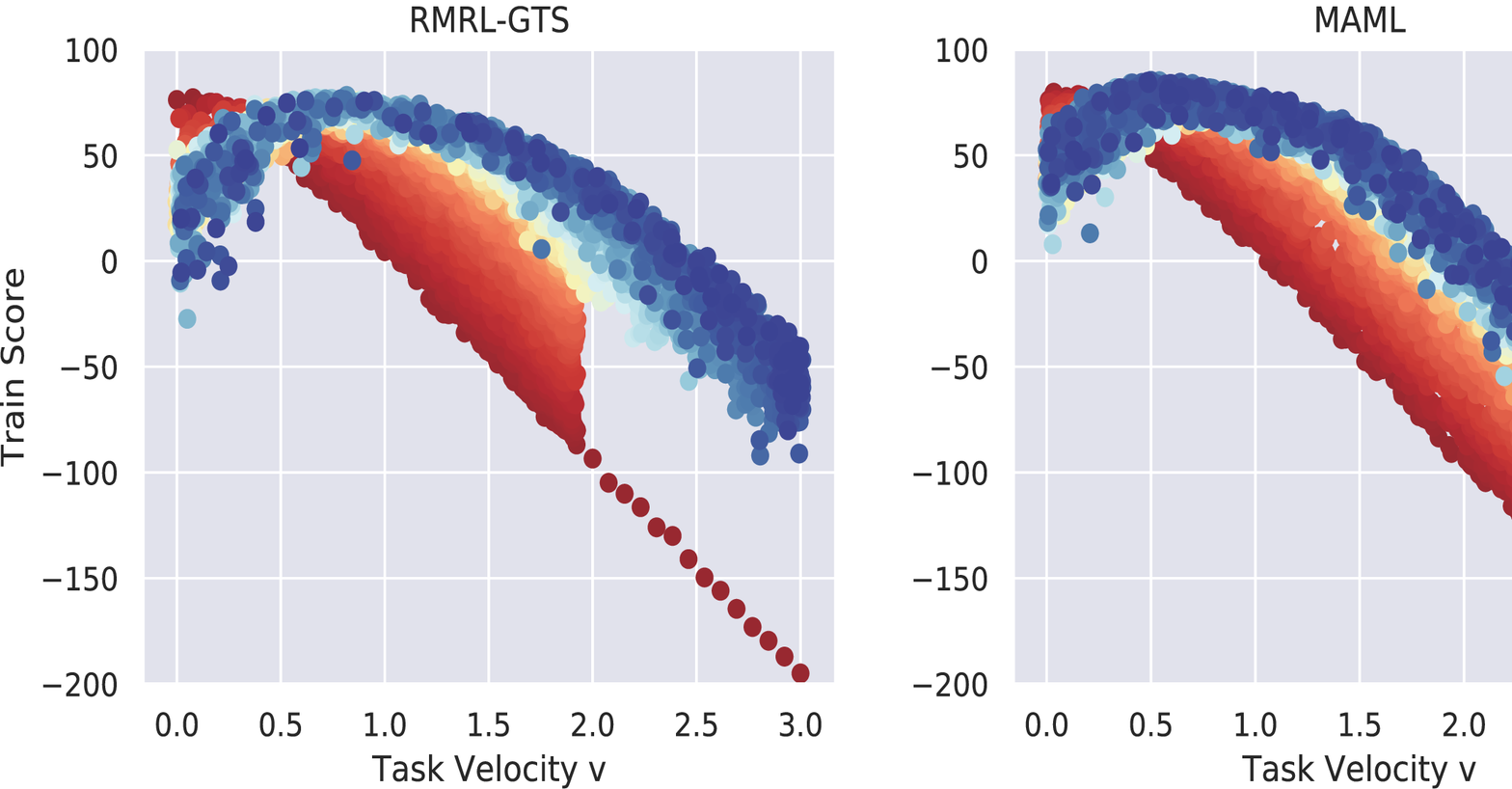}
 \caption{Scores with seed$=1$ during meta-training after adaptation by RMRL-GTS and MAML at $N_{\rm epoch}=200$. } \label{fig:ant_epoch_200_train_sample}
  \end{center}
\end{figure}

Figure \ref{fig:ant_epoch_200_train_sample} show that the reason why scores for difficult tasks acquired by RMRL are higher than MAML 
is due to efficient sampling during meta-training. 
This figure shows the scores after adaptation 
to tasks by RMRL-GTS and MAML with $N_{\rm epoch}=200$ and seed $=1$. 
As shown in the left figure in Fig. \ref{fig:ant_epoch_200_train_sample}, 
except at epoch $=0$, tasks with a large velocity $v\gtrsim 2$ are not 
sampled before epoch $=100$ by RMRL-GTS, the scores for the tasks are gradually 
sampled in the second half of the epoch, and the scores are higher 
than those shown in the right figure (MAML) at a same epoch in Fig. \ref{fig:ant_epoch_200_train_sample}. 
On the other hand, MAML learns tasks with a large velocity from the 
beginning and have a lot of ``useless" sampling. 
These results indicate sampling strategy is important 
when adapting to tasks with a large velocity. 

%%%%%%%%%%%%%%%%%%%%%%%%%%%%%%%%%%%%%%%%%%%%%%%%%%%%%%%%%%%%%%%%%%%%%
%%% Evaluation approach 1, 2 and all in Our Method at N_epoch = 200
%%%%%%%%%%%%%%%%%%%%%%%%%%%%%%%%%%%%%%%%%%%%%%%%%%%%%%%%%%%%%%%%%%%%%
\begin{figure}[!t]
  \begin{center}
 \includegraphics[width=0.6\textwidth, height=0.4\textwidth]{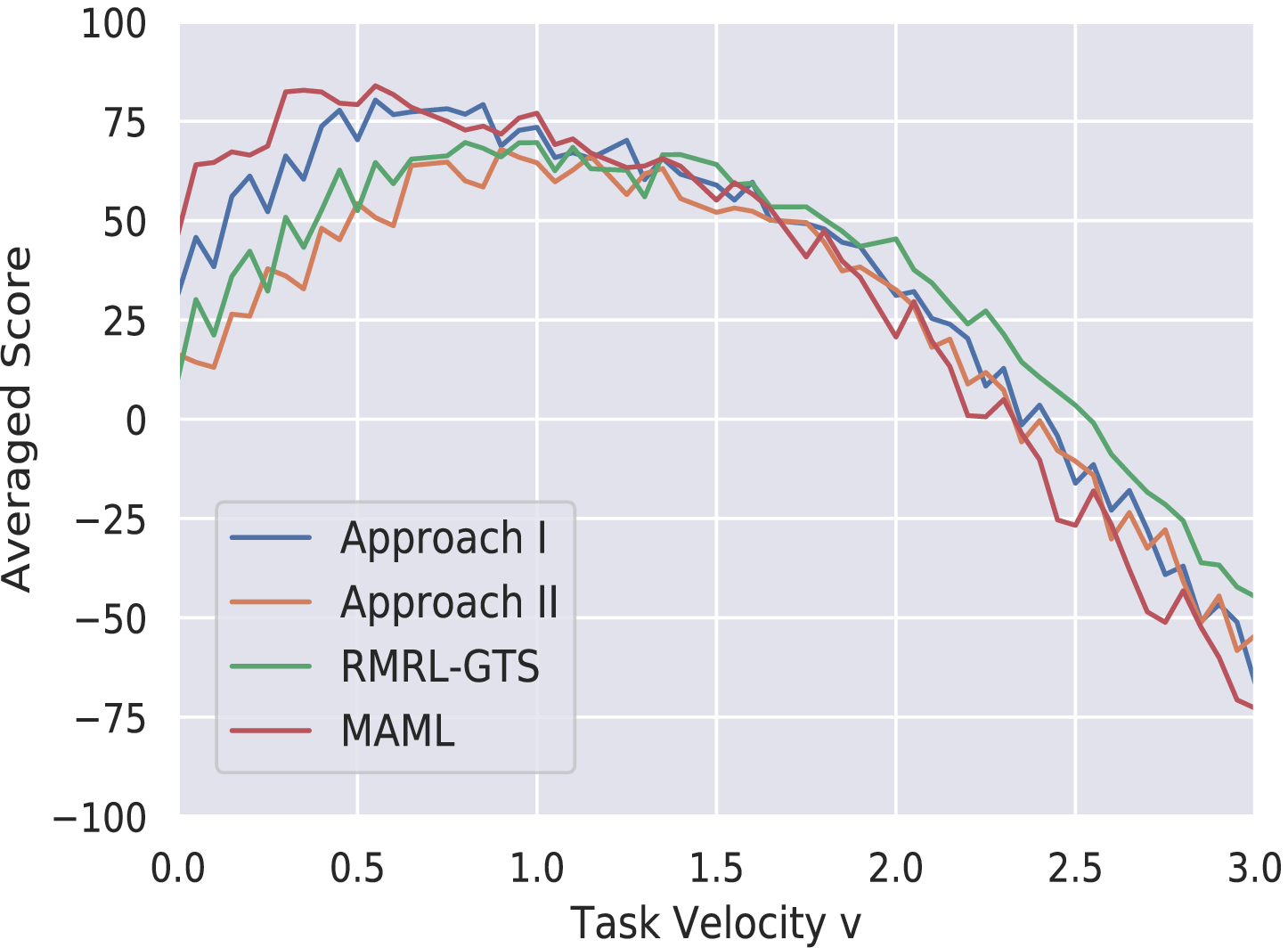}
 \caption{Averaged scores with five seeds for test tasks after adaptation by RMRL-GTS, Approaches I and II, and MAML at $N_{\rm epoch}=200$. } \label{fig:ant_epoch_200_ourapproach}
  \end{center}
\end{figure}

Next, we compared RMRL-GTS with methods using only Approaches I and II as described 
in Sec.  \ref{sec:approach1} and \ref{sec:approach2}, respectively. 
We note that Approach I randomly sampled tasks in $T_{\rm easy}$ and $T_{\rm middle}$, 
and Approach II sampled tasks with the entire task distribution as $T_{\rm easy}$. 
As seen in Fig. \ref{fig:ant_epoch_200_ourapproach}, 
when using the only Approach I, the scores are qualitatively the same as MAML. 
On the other hand, the scores using only Approach II are lower 
than those of RMRL-GTS in most of the task regions. 
Thus, RMRL-GTS (Approaches I and II) is considered to be a robust 
meta-approach compared with using each approach alone, since RMRL-GTS shows the highest scores for high target velocity $v$ 
($v \gtrsim 1.5$). 
In addition, although Approach II 
can adapt to difficult tasks, 
Fig. \ref{fig:ant_epoch_200_ourapproach} shows that robust meta-learning also requires learning 
while restricting the region of sampling tasks as in Approach I. 
Approach I and II are considered to increase sampling efficiency.

%%%%%%%%%%%%%%%%%%%%%%%%%%%%%%%%%%%%%%%%%%%%%%%%%%%%%%%%%%%%%%%%%%%%%
%%% MAML VS Our Method at N_epoch = 200, 400, 600, 800, 1000
%%%%%%%%%%%%%%%%%%%%%%%%%%%%%%%%%%%%%%%%%%%%%%%%%%%%%%%%%%%%%%%%%%%%%

\begin{figure}[!t]
 \includegraphics[width=\textwidth]{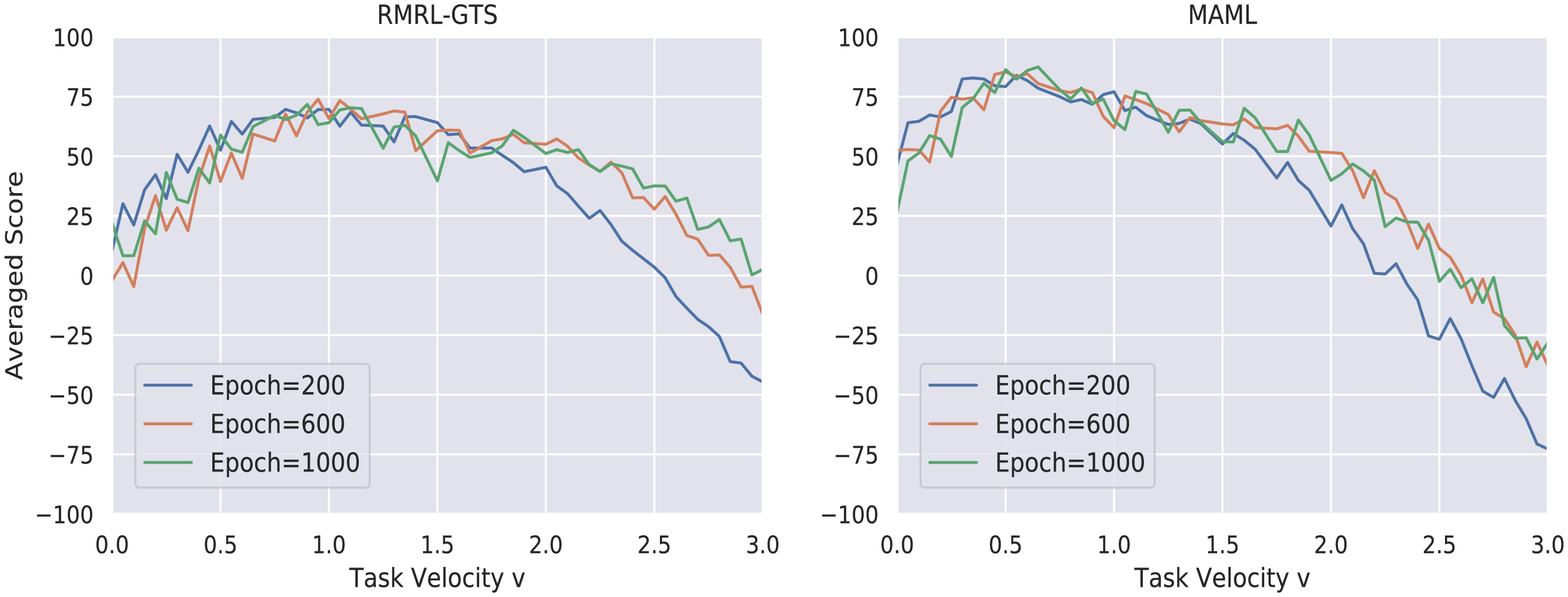}
 \caption{Averaged scores with five seeds for test tasks after adaptation by RMRL-GTS and MAML at $N_{\rm epoch}=200, 600, 1000$. } \label{fig:ant_epoch_200_1000}
\end{figure}

Figure \ref{fig:ant_epoch_200_1000} shows scores at various epochs, 
$N_{\rm epoch}=200, 600, 1000$, by MAML and RMRL-GTS. 
As seen in both figures, both methods tend to show higher scores 
as $N_{\rm epoch}$ is larger for a task, especially for $v\gtrsim 1.5$. 
However, Fig. \ref{fig:ant_epoch_200_1000} shows our method acquires higher scores 
than MAML for difficult tasks ($v\gtrsim 2$). 
On the other hand, scores for easy tasks ($v: [0,1]$) does
not decrease as $N_{\rm epoch}$ increases, which implies 
that our method do not meta-overfit  difficult tasks ($v\gtrsim 2$). 

\begin{figure}[!t]
  \begin{center}
 \includegraphics[width=0.7\textwidth, height=0.35\textwidth]{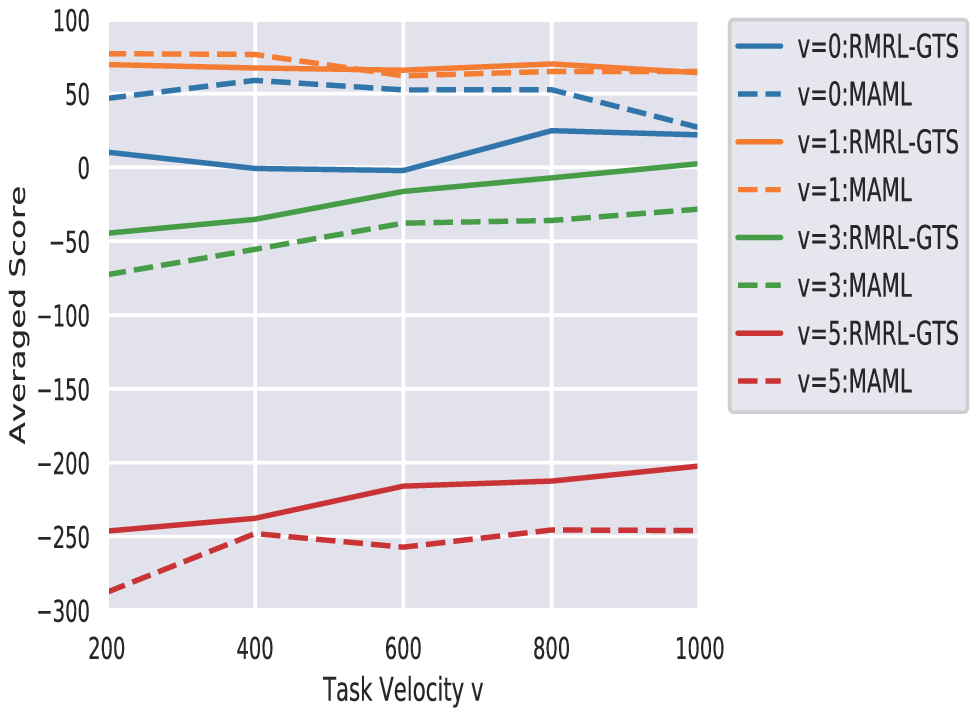}
 \caption{Averaged scores for $N_{\rm epoch}:[200, 1000]$ with five seeds for $v = 0, 1, 3, 5$ learned by RMRL-GTS and MAML. } \label{fig:ant_epoch_200_1000_comparison}
  \end{center}
\end{figure}

Figure \ref{fig:ant_epoch_200_1000_comparison} shows how scores change 
by MAML and RMRL-GTS for specific velocities $v=0, 1,  3, 5$ 
as epoch number $N_{\rm epoch}$ increases. 
Although MAML shows the scores of $v\lesssim 1$ tend to decrease 
as $N_{\rm epoch}$ increases, 
the scores by RMRL-GTS of $v\lesssim 1$ do not change much 
 as $N_{\rm epoch}$ increases. 
As for the scores of the difficult tasks ($v\gtrsim 3$), 
the increase rate of MAML's scores is smaller than that of RMRL-GTS. 
Thus, RMRL-GTS is more robust meta-RL than MAML even when $N_{\rm epoch}$ increases.

\begin{table}[!t]
 \caption{Comparison of results in Ant-Velocity by RMRL-GTS and MAML at $N_{\rm epoch}=1000$. }\label{table:ant}
 \begin{center}
 \begin{tabular}{|l|l|l|}
 \hline
 Evaluation item & RMRL-GTS & MAML\\
 \hline
 highest score & 71.71 & 87.41\\
 $v$ in highest score & 0.90 & 0.65\\
 mean value ($v: [0,3]$) & 44.08 & 44.51 \\
 mean value ($v: [0,5]$) & -15.30 & -33.82 \\
 variance value ($v: [0,3]$) & 354.19 & 1168.51 \\
 bias score at $v=0$ & -49.72 & -60.60\\
 bias score at $v=1$ & -7.65 & -22.49\\
 bias score at $v=2$ & -20.57 & -47.53\\
bias score at $v=3$ & -69.13 & -115.62\\
 ${\rm min} \,v$ with negative reward & 3.05 & 2.50 \\
 \hline
 \end{tabular}
\end{center}
\end{table}

We evaluated results
at 
$N_{\rm epoch}=1000$ shown in Table \ref{table:ant}. 
As previously described, MAML's $v$, where the highest score is 
acquired, is lower than that of RMRL-GTS, 
and  the mean value ($v: [0,5]$) shows RMRL-GTS is more robust  than MAML. 
The variance result shows that RMRL-GTS is less biased, i.e., more robust 
than MAML since the results with MAML are higher than those with our method. 
In addition, we evaluated the bias of scores described as
\begin{equation}
{\rm Bias\, Score} = (\rm{max}_{\tau'}R(\tau') - R(\tau)). \label{eq:result-bias}
\end{equation}
Here, $R(\tau)$ is the mean total reward in Eq. (\ref{eq:reward-total}). 
Eq. (\ref{eq:result-bias}) indicates the difference from the highest score. 
Robust meta-RL approaches for task distribution should require less 
difference between each score and the highest score 
(a high score is also required). 
This is because no matter how high the scores are on particular tasks, 
low scores on other tasks are not considered to be those of robust meta-RL. 
We evaluated robustness for task distribution with Eq. (\ref{eq:result-bias}). 
Table \ref{table:ant} shows all bias scores by RMRL-GTS are higher 
than those by MAML. 
These results also indicate RMRL-GTS is a more robust meta-RL approach than MAML.

\subsection{MuJoCo: HalfCheetah-Velocity}\label{sec:halfcheetah}
\begin{figure}[!t]
 \includegraphics[width=\textwidth]{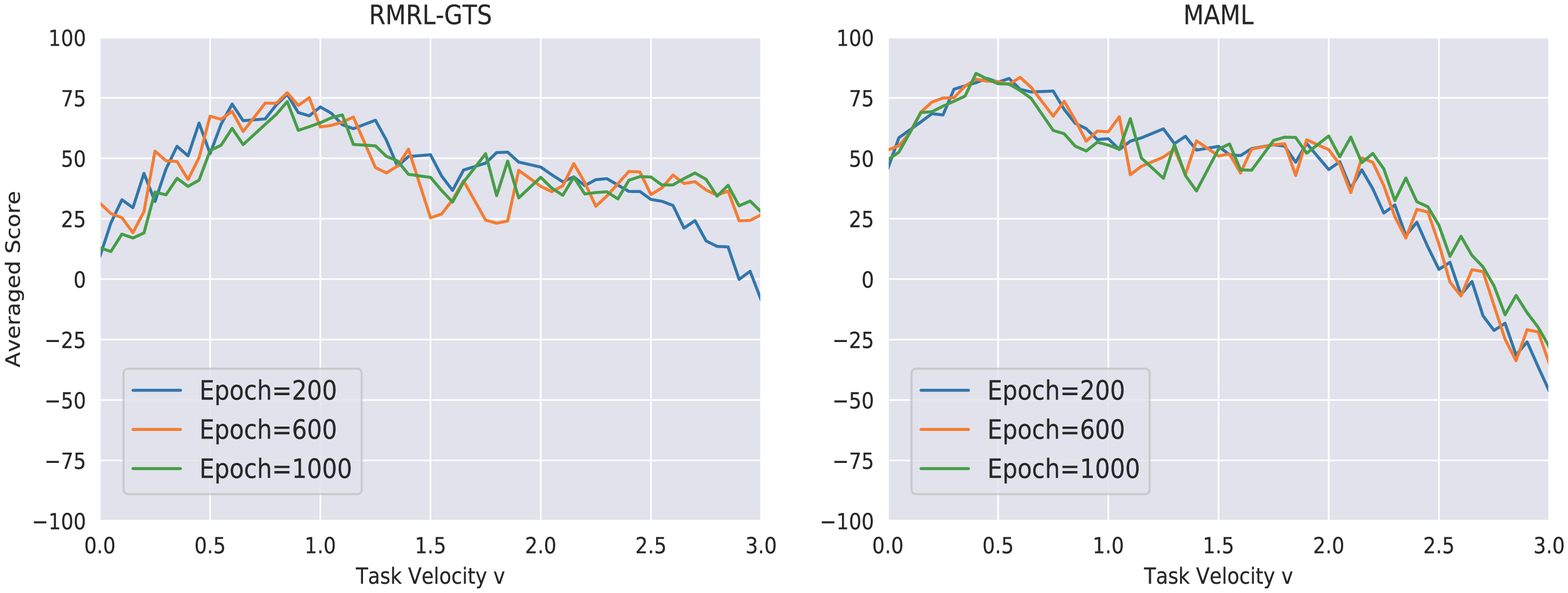}
 \caption{Averaged scores with five seeds for test tasks after adaptation by RMRL-GTS and MAML at $N_{\rm epoch}=200, 600, 1000$. } \label{fig:cheetah_epoch_200_1000}
\end{figure}

In this section, we evaluated RMRL-GTS in the HalfCheetah-Velocity benchmark. 
Figure \ref{fig:cheetah_epoch_200_1000} shows the scores for various epochs, 
$N_{\rm epoch}=200, 600, 1000$. The scores by RMRL-GTS show higher scores 
in the difficult task region ($v\gtrsim 2$) than those by MAML. 
In particular, the scores for $v\gtrsim 2.5$ increases as $N_{\rm epoch}$ increases.

\begin{table}[!t]
 \caption{Comparison of results in HalfCheetah-Velocity by RMRL-GTS and MAML at $N_{\rm epoch}=1000$. }\label{table:halfcheetah}
 \begin{center}
 \begin{tabular}{|l|l|l|}
 \hline
 Evaluation item & RMRL-GTS & MAML\\
 \hline
 highest score & 73.46 & 85.16\\
 $v$ in highest score & 0.85 & 0.40\\
 mean value ($v: [0,3]$) & 42.73 & 45.52 \\
 mean value ($v: [0,5]$) & -6.12 & -27.46 \\
 variance value ($v: [0,3]$) & 195.06 & 748.09 \\
 bias score at $v=0$ & -60.50 & -35.78\\
 bias score at $v=1$ & -8.75 & -29.69\\
 bias score at $v=2$ & -31.31 & -25.92\\
 bias score at $v=3$ & -45.61 & -113.05\\
 ${\rm min} \,v$ with negative reward & 3.30 & 2.75 \\
 \hline
 \end{tabular}
 \end{center}
\end{table}

We evaluated the highest score, mean value, and variance value at 
$N_{\rm epoch}=1000$, and the results are shown in Table \ref{table:halfcheetah}. 
The mean value ($v: [0,5]$) shows RMRL-GTS can acquire higher scores even out of the training task distribution  ($v: [0,3]$) 
than MAML. 
The variance value is better in both methods than that of Ant-Velocity, 
as shown in Table \ref{table:ant}, and that of RMRL-GTS is smaller than that of MAML. 
In addition, RMRL-GTS can adapt to more difficult tasks 
than MAML since the target velocity, at which the score becomes negative 
in our method, is larger than that in MAML. 
These results indicate that RMRL-GTS is a more robust meta-RL approach than MAML even in HalfCheetah-Velocity benchmark. 

\subsection{2D Navigation}
\label{sec:2DNavi}
\begin{figure}[!t]
  \begin{center}
 \includegraphics[width=0.75\textwidth, height=0.5\textwidth]{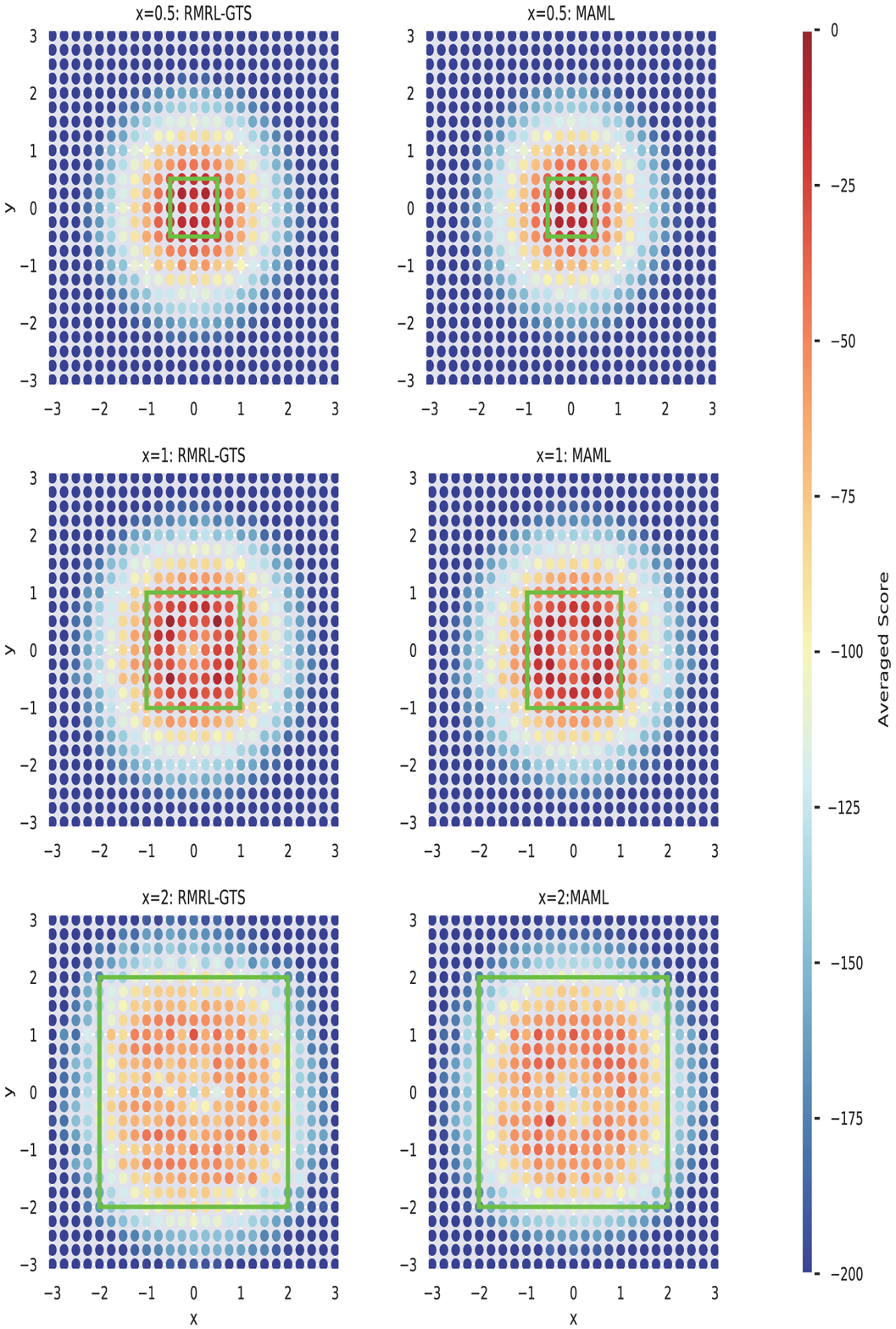}
 \caption{Averaged scores with five seeds for test tasks after adaptation by RMRL-GTS and MAML for 2D Navigation at $N_{\rm epoch}=200$. 
          The region in the green area is meta-training task distribution during meta-learning.} \label{fig:2d_epoch_200}
  \end{center}
\end{figure}

Finally, we evaluated RMRL-GTS in 2D Navigation. 
Figure \ref{fig:2d_epoch_200} shows a comparison of the results 
with RMRL-GTS and MAML in the different train task distributions (green): $|x|, |y|\leq 0.5, 1, 2$. 
These results are evaluated with $|x_{\rm goal}|, |y_{\rm goal}|$ $\leq 3$, 
which is larger than all train task regions considered in this paper. 
At each train task region, although there are no large differences 
in either result compared with those of Ant- and HalfCheetah-Velocity, 
the scores of the edge, near $-200$, by RMRL-GTS 
is slightly larger than that of MAML.

\cite{Mehta} reported that MAML adapted to specific task regions in 2D 
Navigation as the train task region expanded, 
in other words, meta-overfitting. 
However, although we confirmed a small bias, where the highest score 
was not at the center of the task region, i.e. 
$(x_{\rm goal}, y_{\rm goal})=(0,0)$, we could not confirm a large bias as 
seen in \cite{Mehta}  with the same program \cite{Deleu}.

\begin{table}[!t]
 \caption{Comparison of results in 2D-Navigation by RMRL-GTS and MAML at $N_{\rm epoch}=1000$. The region of goal position is $|x_{\rm goal}|, |y_{\rm goal}| \leq 3$. }\label{table:2d}
 \begin{center}
 \begin{tabular}{|l|l|l|}
 \hline
 Evaluation item & RMRL-GTS & MAML\\
 \hline
 mean value ($x=0.5$) & -198.27 & -204.40\\
 mean value ($x=1$) & -169.72 & -173.42\\
 mean value ($x=2$) & -141.67 & -149.85\\
  \hline
  \end{tabular}
  \end{center}
\end{table}

The difference between RMRL-GTS and MAML can be easily seen in Table \ref{table:2d}. 
RMRL-GTS shows a higher mean value than MAML in all train distributions. 
Thus, RMRL-GTS is also a more robust method than MAML even in 2D Navigation, 
which is a wide target task region including negative values 
and a 2D task region.

\section{Conclusion}

We proposed RMRL-GTS, which achieves robust meta-RL for task distribution 
with curricula based on information on scores and task regions. 
Our method acquires a higher score for difficult tasks in the task 
distribution and is less biased in terms of scores in the task distribution 
for meta-RL benchmarks such as Velocity tasks (Ant, HalfCheetah) and 2D Navigation 
than MAML. 
In particular, in Ant-Velocity, our method is superior to 
the previous method \cite{Mehta} for difficult tasks. 
RMRL-GTS shows efficient sampling and the contributions to robust meta-RL. 
In addition to increasing the efficiency of sampling, we show that 
not only an intensive sampling of tasks with poor scores 
but also restricting and expanding the region to be sampled is necessary to 
achieve robust meta-RL (see in Figs. \ref{fig:ant_epoch_200_train_sample} 
and \ref{fig:ant_epoch_200_ourapproach}). 
Our method is not restricted to gradient-based meta-RL such as MAML and 
does not require task information on reward functions, since our method only 
changes the sampling method to improve sampling efficiency. 
Our research will hopefully play an important role in applying RL to 
real-world tasks.

\end{document}